\pdfoutput=1

\documentclass[11pt]{article}

\usepackage[]{emnlp2021}

\usepackage{times}
\usepackage{latexsym}

\usepackage{graphicx}
\usepackage[export]{adjustbox}
\usepackage{caption}
\usepackage[list=false]{subcaption}
\usepackage{comment}
\usepackage[linesnumbered,ruled]{algorithm2e}
\SetKwRepeat{Do}{do}{while}%
\usepackage{amsfonts}
\usepackage{mathtools, cases}

\usepackage{footnote}
\usepackage{hyperref}
\usepackage{tablefootnote}
\usepackage{tabularx}
\usepackage{multicol}
\usepackage{multirow}
\usepackage{tablefootnote}
\usepackage{color, colortbl}
\definecolor{Gray}{gray}{0.9}

\definecolor{cambridgeblue}{rgb}{0.64, 0.76, 0.68}
\definecolor{blue(ncs)}{rgb}{0.0, 0.53, 0.74}
\definecolor{cadetblue}{rgb}{0.37, 0.62, 0.63}
\definecolor{cadmiumgreen}{rgb}{0.0, 0.42, 0.24}
\definecolor{ao(english)}{rgb}{0.0, 0.5, 0.0}
\definecolor{alizarin}{rgb}{0.82, 0.1, 0.26}
\definecolor{vermilion}{rgb}{0.89, 0.26, 0.2}

\usepackage[T1]{fontenc}

\usepackage[utf8]{inputenc}

\usepackage{microtype}

\title{ProtoInfoMax: Prototypical Networks with Mutual Information Maximization for Out-of-Domain Detection}

\author{Iftitahu Ni'mah$^{1,2}$, Meng Fang$^1$, Vlado Menkovski$^1$, Mykola Pechenizkiy$^1$ \\
         Eindhoven University of Technology (TU/e), Eindhoven, The Netherlands$^1$ \\ Research Center for Informatics (BRIN), Bandung, Indonesia$^2$\thanks{$^{*}$Komplek LIPI, Jl. Sangkuriang, Dago, Coblong, Bandung, Indonesia 40135. Phone: (+6222) 2504711.} \\ \texttt{\{i.nimah, m.fang, v.menkovski, m.pechenizkiy\}@tue.nl}$^1$}

\begin{document}
\maketitle
\begin{abstract}
The ability to detect Out-of-Domain (OOD) inputs has been a critical requirement in many real-world NLP applications. For example, intent classification in dialogue systems. The reason is that the inclusion of unsupported OOD inputs may lead to catastrophic failure of systems. However, it remains an empirical question whether current methods can tackle such problems reliably in a realistic scenario where zero OOD training data is available. In this study, we propose ProtoInfoMax, a new architecture that extends Prototypical Networks to simultaneously process in-domain and OOD sentences via Mutual Information Maximization (InfoMax) objective. Experimental results show that our proposed method can substantially improve performance up to 20\% for OOD detection in low resource settings of text classification. We also show that ProtoInfoMax is less prone to typical overconfidence errors of Neural Networks, leading to more reliable prediction results. \footnote{Code and preprocessed data are available at \href{https://github.com/inimah/protoinfomax}{https://github.com/inimah/protoinfomax.git}.} 

\end{abstract}

\section{Introduction}
\label{sec:intro}
\begin{figure}[!t]
    \centering
    \includegraphics[width=.45\textwidth]{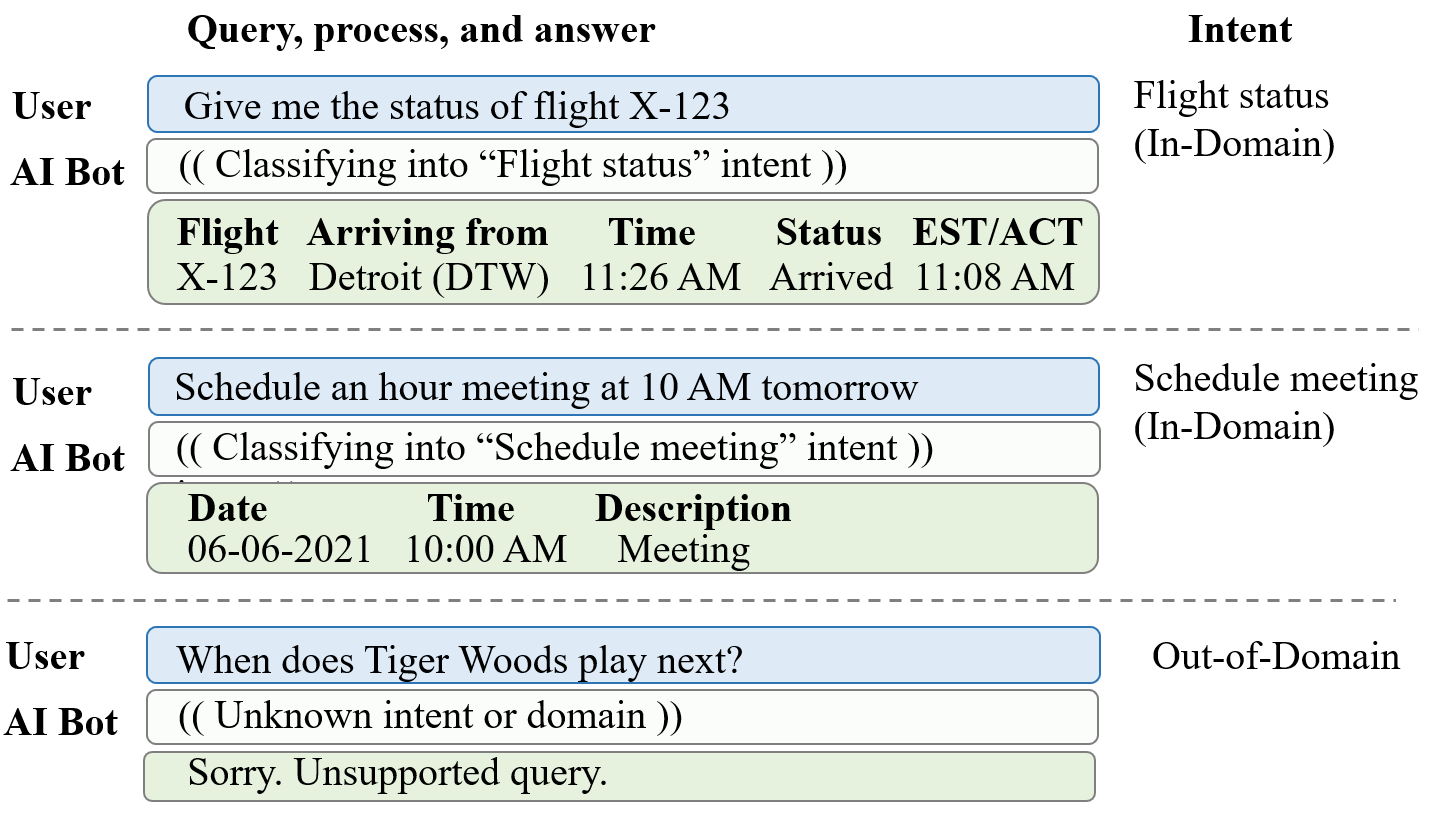}
    \caption{An example of OOD detection in task-oriented dialog systems.}
    \label{fig:ood_intent_exp}
    \vspace{-1em}
\end{figure}

Many real-world applications imply an open world assumption \cite{6365193,fei-liu-2016-breaking} \footnote{System
built under this assumption should be able to not only correctly analyze In-Domain (ID) inputs but also reliably reject Out-of-Domain (OOD) inputs that are not supported by the system.}, requiring intelligent systems to be aware of novel Out-of-Domain (OOD) examples, given limited In-Domain (ID) and zero OOD training data. Intent classification for conversational AI services, for instance, may have to deal with unseen OOD utterances \cite{tan-etal-2019-domain,kim2018joint,larson-etal-2019-evaluation,zheng2020out}. Question answering system is
 also preferred to have a certain degree of language understanding via its ability to contrast between relevant and irrelevant sentences \cite{yeh-chen-2019-qainfomax}. Likewise, a classifier trained on past topics of social media posts is often expected to be aware of future social media streams with new unseen topics \cite{fei-liu-2016-breaking,Fei2016LearningCumulati}. An example of an AI system with OOD awareness is illustrated in Figure~\ref{fig:ood_intent_exp}. When a user inputs an unknown query with OOD intent, instead of providing random feedback, a system that is aware of OOD inputs can better respond informatively.

To develop a reliable intelligent system that can correctly process ID inputs and detect unclassified inputs from different distribution (OOD), existing approaches often formulate OOD detection as anomaly detection \cite{RYU201726,ryu-etal-2018-domain, hendrycks2019oe}. The concept of learning ID classification and OOD detection tasks simultaneously is also incorporated in diverse applications, including open text classification \cite{shu-etal-2017-doc} and OOD detection in task-oriented dialog system \cite{Kim2018,zheng2020out}. These methods rely on large-scale ID and OOD labeled training data or well-defined data distributions.

Unfortunately, large data settings make the methods unrealistic for many real world applications with limited ID and zero OOD training data.
As a result, current research introduces few-shot and zero-shot learning frameworks for OOD detection problems in a low resource scenario of text classification \cite{tan-etal-2019-domain}. Their objective is to learn a metric space for ID and OOD prediction given prototype representation of ID sentences and target sentences sampled from ID and OOD distribution. However, the current method neglects an overconfidence issue of the trained Prototypical Networks in the inference stage where both novel ID and OOD inputs occur. For example, OOD samples are likely to be classified as ID with a high similarity score \cite{liang2018enhancing,Shafaei2019}, especially if they share common patterns or semantics with ID samples (e.g. common phrases, sentence topicality, sentiment polarity) \cite{lewis2018generative}.

To mitigate the above problems, we adopt Mutual Information Maximization (InfoMax) objective \cite{belghazi2018mutual,hjelm2019learning} for regularizing Prototypical Networks (Section~\ref{sec:infomax_reg}). We extend Prototypical Networks \cite{snell2017prototypical} to learn multiple prototype representations by maximizing Mutual Information (MI) estimates between sentences that share a relevant context, such as keywords (Section~\ref{sec:proto-gen}). We demonstrate that our proposed method is less prone to typical overconfidence error of Neural Networks \cite{Lakshminarayanan2017Simple,Guo2017Calibration,liang2018enhancing,Shafaei2019}. This result leads to more reliable prediction outcomes, specifically in the inference stage where the model has to deal with both novel ID and OOD examples. Overall, experimental results on real-world low-resource sentiment and intent classification (Section~\ref{sec:results}) show that the proposed method can substantially improve performance of the existing approach up to 20\%.

To summarize, our contributions are as follows:
 
\begin{itemize}
    \item We introduce ProtoInfoMax -- Prototypical Networks that learn to distinguish between ID and OOD representations via Mutual Information Maximization (InfoMax) objective (Section~\ref{sec:infomax_reg}). 
  
    \item We enhance ProtoInfoMax by incorporating multiple prototype representations (Section~\ref{sec:proto-gen}) to further improve the discriminability of the learned metric space.

\item We further investigate the reliability of Prototypical Networks in this study, in addition to common metrics used for evaluating Out-of-Domain detection (Section~\ref{sec:rel-id}-\ref{sec:rel-ood}). Our problem of interest is determining whether the learned metric space indicates a well calibrated model. That is, a condition where the trained model assigns high similarity score for test samples drawn from ID distribution and assigns lower similarity score for samples drawn from OOD distribution.
    
\end{itemize}

\section{Related Work}
\label{sec:related_work}

\subsection{Few-shot Learning}
Few-shot Learning (FSL) has been increasingly studied in NLP. Several works have adopted the experimental protocol of FSL, expanding the application of FSL in 
text classification \cite{yu-etal-2018-diverse, bao2019few, tan-etal-2019-domain,zhang-etal-2020-discriminative} and other tasks \cite{fang2017learning,han-etal-2018-fewrel,gao-etal-2019-fewrel,sun-etal-2019-hierarchical,dopierre-etal-2021-neural}.

\subsection{Out-of-Domain Detection}
The problem of Out-of-Domain (OOD) detection has been investigated in many contexts; such as anomaly detection \cite{zenati2018adversarially,hendrycks2019oe}, one-class classification \cite{khan2014one,pmlr-v80-ruff18a}, open-set recognition \cite{geng2020recent}, and novelty detection \cite{perera2019ocgan}. In speech recognition and language understanding domain, the problem is formulated as OOD utterances detection \cite{lane2006out,tur2014detecting,zheng2020out}. Most of these works, however, depend on the availability of large-scale ID and OOD samples, in addition to the inclusion of OOD samples in training data as supervision signals for the model.


\subsection{Mutual Information Objective}

Incorporating Maximization Mutual Information (MMI) \cite{Linsker1988Self,Bell1995Information}, which we refer to as InfoMax (section~\ref{sec:infomax_reg}), as training objective for Neural Networks is exemplified by early work on diversifying neural conversational model \cite{li-etal-2016-diversity}. However, the work mainly uses the MMI objective in the inference stage for controlling the decoder outputs. Recent works on InfoMax objective for deep learning \cite{belghazi2018mutual,hjelm2019learning} have introduced simple yet effective loss function approximation, such that the objective can be used in the training stage. Prior to our study, InfoMax objective has been adapted to learn useful representations by maximizing relevant information between local and global features of image data \cite{hjelm2019learning}, to learn speaker representations \cite{ravanelli2019learning}, and to learn robust question answering system \cite{yeh-chen-2019-qainfomax}.


\begin{figure}[!ht]
    \centering
    \includegraphics[width=.45\textwidth]{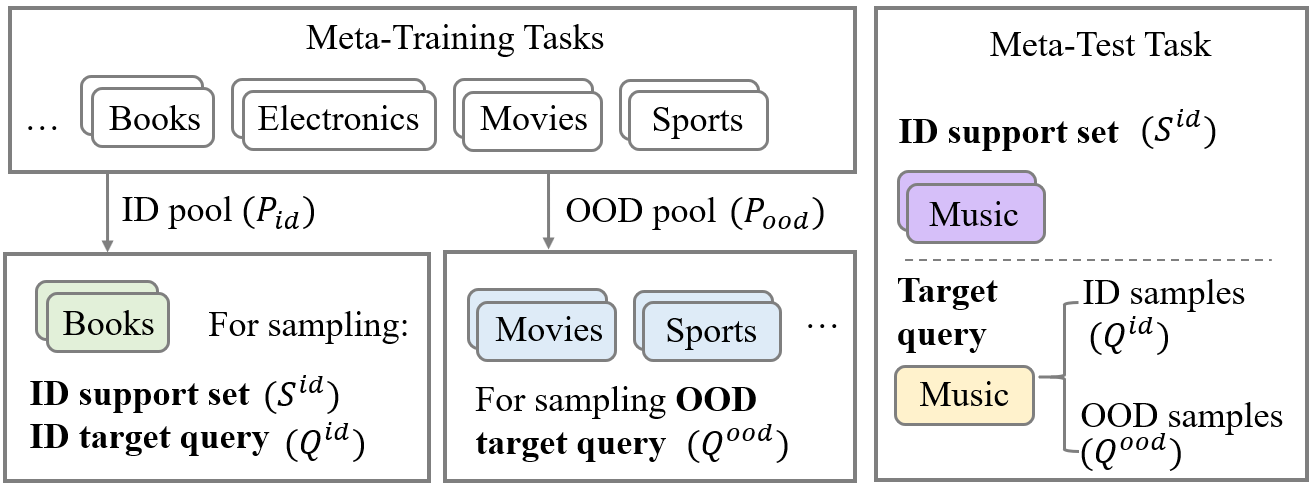}
    \caption{Illustration on how ID and OOD sentences are sampled during meta-training and meta-test tasks. (``Books'', ``Electronics'', \ldots) exemplifies domains available in training data, disjoint with examples in test data. 
    }
    \label{fig:meta-tr}
    \vspace{-1em} 
\end{figure}

\section{Problem Definition}
\label{sec:problem_def}

\begin{figure*}[!ht]
    \centering
    \begin{subfigure}[t]{.475\textwidth}
         \centering
         \includegraphics[width=\linewidth]{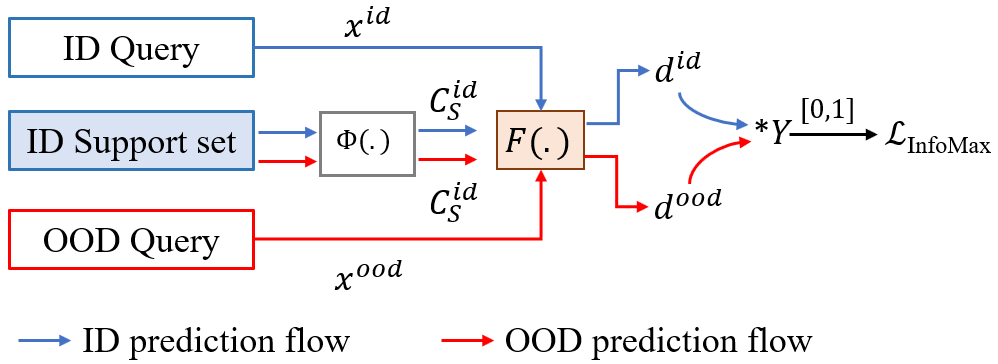}
         \caption{}
	\label{fig:imax-flo}
       \end{subfigure}
       \hfill
    \begin{subfigure}[t]{.425\textwidth}
         \centering
         \includegraphics[width=\linewidth]{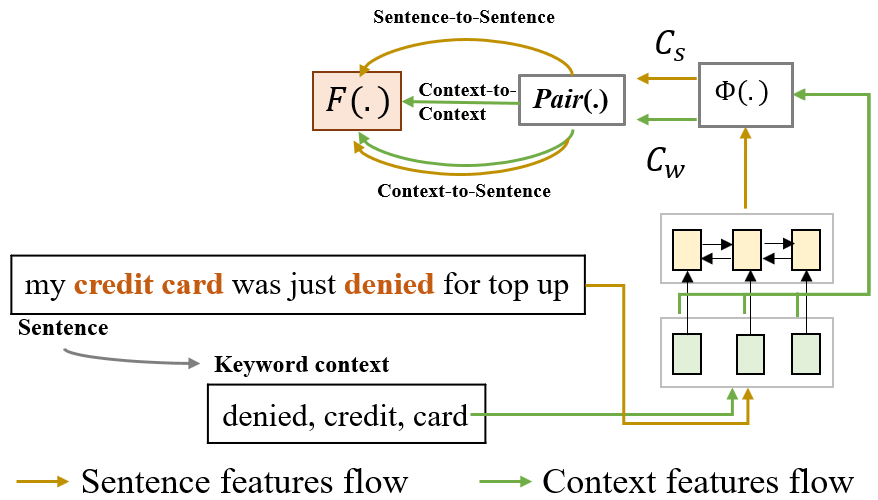}
         \caption{}
	\label{fig:imax-flo+}
       \end{subfigure}
    \caption{Proposed ProtoInfoMax. (a) ProtoInfoMax with prototype vector based on sentence features $C_S$. The encoder architecture that projects raw inputs into vectors in metric space is omitted to simplify the illustration. $C_S^{id}$ is drew as separated ID and OOD lines to help the illustration, but in reality there is only one ID support set that is used to compute similarity scores $d^{id}$ and $d^{ood}$. (b) A partial illustration of ProtoInfoMax++ with multiple prototype vectors ($C_S , C_W$), correspond to prototype vector based on sentence features and keyword context features respectively. Green boxes represent word embeddings in the encoder module of Prototypical Networks. Yellow boxes represent bidirectional GRU as sentence embedding layer.}
    \label{fig:prototinfomax}
\end{figure*}

Similar to the previous setting \cite{tan-etal-2019-domain}, we consider the zero-shot OOD detection problem for meta-tasks in this study. In general, there are three main inputs for prototypical learning in this study: ID support set $S^{id}$, ID target query $Q^{id}$, and OOD target query $Q^{ood}$ (Figure~\ref{fig:meta-tr} and Figure~\ref{fig:imax-flo}).

\paragraph{Meta-training tasks}


For each training episode, ID examples $\mathcal{D}_{id}$ are drawn independently from ID distribution in meta-training tasks $\mathcal{P}_{T_i}, T_i\in\mathcal{T}$ (later refer to as $\mathcal{P}_{id}$). Specifically:

\vspace{-1em}
\begin{equation}
    \mathcal{D}_{id} = \{(x_1,y_1), \ldots, (x_n,y_n)\} \sim \mathcal{P}_{T_i},
   \vspace{-.5em}
\end{equation}

\noindent where each ID example is composed of sentences $x$ and their corresponding ID classes $y$. In sentiment classification benchmark, this $y$ is a sentiment label, $y \in $ \{``positive'' or ``negative''\}. In Figure~\ref{fig:meta-tr}, $\mathcal{P}_{T_i}$ or $\mathcal{P}_{id}$ is described as ID domain ``Books''. ID support set $S^{id}$ and ID target query $Q^{id}$ are drawn from $\mathcal{D}_{id}$, where $S^{id}$ and $Q^{id}$ are mutually exclusive: $S^{id} \cap Q^{id} = \emptyset$. 


OOD data $\mathcal{D}_{ood}$ is drawn from out-of-episode distribution $\mathcal{P}_{\mathcal{T}_j}$:

\vspace{-1em}
\begin{equation}
    \mathcal{D}_{ood} = \{(x_1,y_1), \ldots, (x_n,y_n)\} \sim \mathcal{P}_{T_j}, 
\vspace{-.5em}
\end{equation}

\noindent where tasks or domains $\mathcal{T}_j$ are disjoint with those in training: $\mathcal{T}_j \in \mathcal{T}, \mathcal{T}_j \neq \mathcal{T}_i$ (later refer to as $\mathcal{P}_{ood}$). $\mathcal{P}_{\mathcal{T}_j}$ or $\mathcal{P}_{ood}$ is described as out-of-scope domains (``Movies'', ``Sports'', $\ldots$) in Figure~\ref{fig:meta-tr}.

\paragraph{Meta-test tasks} As illustrated in Figure~\ref{fig:meta-tr}, each task or domain in the test set (e.g. ``Music'') is composed of a disjoint ID support set and target queries (ID and OOD). ID examples are accompanied with ID class labels, i.e. $y \in$ \{``positive'', ``negative''\} for sentiment classification benchmark, while OOD examples are annotated with OOD labels ($y=$``ood''). Note that this $y=$``ood'' is unknown to the model during training.



\section{ProtoInfoMax}
\label{sec:proto}

We propose two models: \textbf{ProtoInfoMax} and \textbf{ProtoInfoMax++}, briefly illustrated in Figure~\ref{fig:prototinfomax}. The main difference between the two models is their prototype generator, further discussed in Section~\ref{sec:proto-gen}. \textbf{ProtoInfoMax++} merges multiple prototype representations: (i) standard feature averaging prototype vector based on sentence features, referred to as $C_{S}$; and (ii) prototype vector based on keyword context features, referred to as $C_{w}$. We regularize both models with an InfoMax objective, discussed in Section~\ref{sec:infomax_reg}. 


\subsection{InfoMax Objective}
\label{sec:infomax_reg}

We adopt the recently proposed Mutual Information Maximization (InfoMax) training objective for deep learning \cite{belghazi2018mutual,hjelm2019learning} as a contrastive view of data drawn from ID and OOD distribution. The idea is simple: we want to maximize Mutual Information (MI) estimates for samples drawn from ID distribution $\mathcal{P}_{id}$, while penalizing OOD samples with lower MI estimates.



Here, a multi-objective for simultaneously classifying ID sentences and detecting OOD sentences is formulated as a contrastive learning framework via an InfoMax objective. The model is enforced to learn binary reject function $\mathcal{L}$ that partitions the input space $\mathcal{X}$ with respect to $\mathcal{P}_{\textnormal{id}}$ and $\mathcal{P}_{\textnormal{ood}}$. Incorporating a binary reject function for regularizing Prototypical Networks in the current OOD detection problem can simplify the overall training mechanism. 
Namely, it can be approximated by a simple cross-entropy (BCE) loss implementation of InfoMax objective \cite{hjelm2019learning,yeh-chen-2019-qainfomax} \footnote{For a theoretical justification on how binary cross-entropy (BCE) loss approximates Mutual Information (MI) between two random variables, including the alternatives, we refer reader to the prior works on investigating InfoMax objective for deep representation learning \cite{belghazi2018mutual,hjelm2019learning,48397,Lingpeng2020MutualInformation}.}. In the current OOD detection problem, the loss is formulated as the approximation of MI between prototype vector of ID support set and target queries $I(C^{id}, Q)$:


\vspace{-1em}
\begin{equation}
    \begin{aligned}
    I(C^{id}, Q) \geq \mathbb{E_P}[\log F(C^{id}, x^{id})] +\\
    \mathbb{E_Q}[\log (1-F(C^{id}, x^{ood}))],
    \end{aligned}
  \vspace{-.5em}
\end{equation}

\noindent where $\mathbb{E_P}$ and $\mathbb{E_Q}$ denote the expectation over ID and OOD samples respectively. $x^{id}$ and $x^{ood}$ are ID and OOD examples as target queries, $x^{id} \in Q^{id}$, $x^{ood} \in Q^{ood}$, $\{Q^{id}, Q^{ood}\} \in Q$. $F(.)$ is a similarity scoring layer.


InfoMax loss is then defined as binary cross-entropy loss between ID and OOD prediction:


\vspace{-.5em}
\begin{equation}
    \begin{aligned}
    \mathcal{L}(C^{id},Q)  = \frac{1}{|Q^{id}|} \sum_{x^{id} \in Q^{id}} \log F(C^{id}, x^{id}) +\\
    \frac{1}{|Q^{ood}|} \sum_{x^{ood} \in Q^{ood}} \log (1-F(C^{id}, x^{ood})).
    \end{aligned}
   \vspace{-.5em}
\end{equation}

\subsection{Learning Framework}
\label{sec:framework}

Figure~\ref{fig:imax-flo} illustrates the proposed model with an InfoMax objective. The prediction outcome is represented by similarity scores between class prototypes and target queries, resulting in scores for ID targets $(d^{id})$ and OOD targets $(d^{ood})$. Since the training objective mainly focuses on promoting the separability between ID and OOD representations, we preserve ID supervision signals, i.e. $y \in$ \{``positive'', ``negative''\} in sentiment classification benchmark, by projecting similarity scores ($d^{id}, d^{ood}$) into representation space of $y$. Thus, the final prediction is defined as: $\hat{y}=d * Y; d=\{d^{id}, d^{ood}\}; Y\in \mathbb{R}^{b \times d}$.




\subsection{Prototype Generator $\Phi(.)$}
\label{sec:proto-gen}

For both proposed models, we use standard prototype generator $\Phi(.)$ based on feature averaging. 
Given encoded representations of ID support set $S^{id} \in \mathbb{R}^{b \times k \times d}$ ($b=$ batch size, $k=$ number of examples in support set, $d=$ dimension size of output representations), the prototype vector $C^{id}$ is described as an averaged representation of those $k$-representations: $C^{id} = \Phi(S^{id}_k)= \frac{1}{k}\sum_{i=1}^k S^{id}_{i}$.

\paragraph{Sentence-based Features}

Given encoded representations of sentences in ID support set $S^{id}$, class prototype vector $C^{id}$ is defined as a mean vector of those sentence features: $C^{id} = \Phi(S^{id}_k)$. To prevent confusion, prototype vector based on sentence features is later denoted as $C_S$. 


\paragraph{Keyword-based Features}

In an extremely low resource setting where training data may provide insufficient contexts due to the scarcity of novel sentences, the model may not be able to learn meaningful sentence representations. To better guide the learning, we utilize keywords as auxiliary inputs for ProtoInfoMax++ (Figure~\ref{fig:imax-flo+}). 



Intuitively, sentences drawn from the same domain or intent distribution may share relevant context via their keywords. Therefore, keywords can be viewed as local context representation of a sentence. The more keywords that two sentences share in common, the more similar or related the two sentences are. While, from the perspective of word orientation in embedding space, keywords that are close together with respect to their angular distance are expected to carry similar semantic meaning. Sentences containing those similar subset of keywords can be considered to carry similar or related semantics. This motivates us to incorporate keyword representations into the current prototypical learning problem.



Prototype vector $C_w$ is defined as a mean vector representation of sentence's keywords $W=\{w_1, w_2, \ldots, w_n\}$ weighted by their corresponding Idf value: 



\vspace{-1em}
\begin{equation}
    \begin{aligned}
    C_w = \frac{1}{n} \sum_{i=1}^n (w_{i}*Idf_{i}).
    \end{aligned}
    \vspace{-.5em}
\end{equation}

\noindent Since sentence inputs are composed of ID support set $S^{id}$, ID target queries $Q^{id}$ and OOD target queries $Q^{ood}$, prototype vector based on keyword features can be further denoted as: $C_w^{sup}$ representing keywords in ID support set; and $C_w^Q$ representing keywords in target queries. For ID support set containing $k$-sentences, $C_w$ is averaged over $n$-keywords and $k$-sentence features: $C_w^{sup} =\frac{1}{k} \frac{1}{n} \sum_{j=1}^k \sum_{i=1}^n (w_{i}*Idf_{i})^{j}$. 


\begin{table*}[!ht]
\resizebox{\textwidth}{!}{
    \centering
    \begin{tabular}{l|cll|ccc|ccc}
    Data & \multicolumn{3}{c|}{Meta-training} & \multicolumn{3}{c|}{Meta-validation} & \multicolumn{3}{c}{Meta-test} \\
    & \#Task & \#ID class & \#Sample & \#Task & \#ID sample & \#OOD sample & \#Task & \#ID sample & \#OOD sample \\
    \hline
    Amazon-rev (Sentiment)     & 13 & 2 (shared) & 2M/task & 4 & (200,50) & 20 & 4 & (200,50) & 20\\
    AI-conv (Intent) & 4 & 10 (disjoint) & 100-3K/ID class & 10 &  (120,30)& 20 & 10 & (120,30) & 20\\
    \hline
    \end{tabular}
}
    \caption{Data set statistics. \#ID sample in meta-validation and meta-test $(n_A,n_B)$ shows the number of disjoint samples for ID support set and total samples for ID target query respectively. Except for meta-training in AI-conv data (Intent), number of samples (\#Sample, \#ID sample, \#OOD sample) are shown as a figure representing examples within each task or domain.   
    }
    \label{tab:data_stats}
\end{table*}

\subsection{Similarity function $F(.)$}
\label{sec:multi-sims}

For model utilizing keyword auxiliary inputs (ProtoInfoMax++), we use a multi-perspective of similarity function $F(.)$ to calculate similarity score $d$ between support set and target queries.

\paragraph{Sentence-to-sentence similarity $F(C_S, Q)$}
    
    -- This function is by default similarity measure for all Prototypical Networks in this study. Here, $C_S$ denotes prototype vector of ID support set and $Q$ is sentence embedding projection of target queries.

\paragraph{Context-to-context similarity $F(C_w^{\textnormal{sup}},C_w^{\textnormal{Q}})$}
    
    -- We want to maximize MI between prototype representation of keywords in support set and target queries. $C_w^{\textnormal{sup}}$ is prototype vector computed from keyword contexts in support set, while $C_w^{\textnormal{Q}}$ is computed from keywords in target queries. 
  
\paragraph{Context-to-sentence similarity $F(C_w^{\textnormal{sup}*},C_w^{\textnormal{Q}*})$}
    
    -- We want to maximize MI between sentences that share relevant context or keyword representations. Sentence representations with respect to their keyword contexts are computed as an element-wise matrix multiplication between encoded sentences and encoded keywords: $C_w^{\textnormal{sup}*} = C_w^{\textnormal{sup}} * C_S$; $C_w^{\textnormal{Q}*} = C_w^{\textnormal{Q}} * Q$.

\subsection{Total Loss}

\paragraph{ProtoInfoMax}

Given prototype vector based on sentence features $C_S$ and target queries $Q$ drawn from $\mathcal{P}_{id}$ and $\mathcal{P}_{ood}$, the loss function for ProtoInfoMax is described as error loss given prototype vector generated from sentence features $C_S$ and target queries $Q$:

\begin{equation}
    \begin{aligned}
    \mathcal{L}_{\small \texttt{\textbf{infomax}}} = \mathcal{L}(C_S, Q).
    \end{aligned}
\end{equation}

\paragraph{ProtoInfoMax++}

The total loss for ProtoInfoMax++ is described as cumulative losses given sentence-to-sentence similarity $F(C_S, Q)$, context-to-context similarity $F(C_w^{\textnormal{sup}},C_w^{\textnormal{Q}})$, and context-to-sentence similarity $F(C_w^{\textnormal{sup}*},C_w^{\textnormal{Q}*})$ (section ~\ref{sec:multi-sims}):
\begin{equation}
    \begin{aligned}
    \mathcal{L}_{\small \texttt{\textbf{infomax++}}} = \mathcal{L}(C_S, Q) + \\ \mathcal{L}(C_w^{\textnormal{sup}},C_w^{\textnormal{Q}}) + \\ \mathcal{L}(C_w^{\textnormal{sup}*},C_w^{\textnormal{Q}*}).
    \end{aligned}
\end{equation}



\section{Experiments}
\label{sec:exp}

\subsection{Dataset}
\label{sec:dataset}

\paragraph{Amazon Product Reviews}
For structuring Amazon review data into meta-tasks, we followed strategy from previous works on few-shot classification \cite{yu-etal-2018-diverse,tan-etal-2019-domain}. 


\paragraph{AI Conversational Data} For constructing intent classification meta-tasks, we use two data sets that share contexts: AI Conversational Data \cite{chatterjee-sengupta-2020-intent}; and (CLINC150) \cite{larson-etal-2019-evaluation,casanueva-etal-2020-efficient} \footnote{We use different benchmarks for intent classification task because the footage of preprocessed data from previous work \cite{tan-etal-2019-domain} is unavailable publicly.}. The preprocessed data contains disjoint classes across tasks, introducing a more challenging ID and OOD prediction task for Prototypical Networks in this study. In meta-training, each task (domain) is composed of 10 intent category labels ($N=10$). Meta-validation and meta-testing are constructed from CLINC150. We use $N=1$ and $N=2$ set up to inspect model performance on one ID class and multiple ID classes prediction respectively. 


\subsection{Model and Hyper-parameters}
\label{sec:model_hyperparams}
\paragraph{Baselines} We use two baselines: 1) \textbf{Proto-Net} \cite{snell2017prototypical,yu-etal-2018-diverse}, a native Prototypical Network with entropy-based loss function; 2) \textbf{O-Proto} \cite{tan-etal-2019-domain}, state-of-the-art approach for simultaneously learning ID classification and OOD detection. We do not include previous approaches based on non-Prototypical Networks (OSVM \cite{Bernhard2001}, LSTM Autoencoder \cite{RYU201726}, and vanilla CNN \cite{tan-etal-2019-domain}) because these methods were shown to be under-performed in \cite{tan-etal-2019-domain}. We want to focus on further inspecting the reliability aspect of simple Prototypical Networks without additional learning pipelines.

\paragraph{Hyper-parameters} 
For all models, we initialized word representation from pretrained fastText \footnote{https://fasttext.cc/}. We updated fastText representation by further training it on current benchmark data, before using it as initialization for word embedding layer of Prototypical Networks. We used Tf-Idf \cite{sparck1972statistical,salton1988term} as a keyword extraction method in the preparation of auxiliary inputs (keywords) for ProtoInfoMax++ due to its simple assumption. Namely, TfIdf measures word importance based on co-occurrence of words within a small group of documents. For future reference, this TfIdf approach can be substituted by any automated keyword extraction methods.

We use one layer Bidirectional-GRU (dimension size=200) as backbone encoder architecture for all models; and one layer Attention Network that is initialized based on $r$ context query representations ($r=5$) sampled from uniform distribution $\mathcal{U}[.1,.1]$. Similarity scoring layer $F(.)$ is based on cosine similarity via matrix multiplication between prototype vector and target queries. All models were trained up to 60 epochs with batch size 100. Note that each epoch contains \#Tasks that are dynamically sampled as training episodes. 






\begin{table*}[!ht]
\resizebox{\textwidth}{!}{
    \centering
    \begin{tabular}{ | l | cc | c c |cc | cc |  c | c |c |}
    \hline
    Method & \multicolumn{6}{c|}{Sentiment Cls ($N=2$)} & \multicolumn{2}{c|}{Intent Cls ($N=1$)} & \multicolumn{3}{c|}{Intent Cls ($N=2$)}\\
    & \multicolumn{2}{c|}{EER} & \multicolumn{2}{c|}{CER$^{\textnormal{id}}$} & \multicolumn{2}{c|}{CER$^{\textnormal{all}}$}  & \multicolumn{2}{c|}{EER} & 
    \multicolumn{1}{c|}{EER} & \multicolumn{1}{c|}{CER$^{\textnormal{id}}$} & \multicolumn{1}{c|}{CER$^{\textnormal{all}}$}  \\ 
    & $\mathcal{T}^{\textnormal{val}}$ & $\mathcal{T}^{\textnormal{test}}$ & $\mathcal{T}^{\textnormal{val}}$ & $\mathcal{T}^{\textnormal{test}}$ & $\mathcal{T}^{\textnormal{val}}$ & $\mathcal{T}^{\textnormal{test}}$ & $\mathcal{T}^{\textnormal{val}}$ & $\mathcal{T}^{\textnormal{test}}$  &&&\\
    \hline
    \bf \underline{Baselines} & && & &&  && &  &&\\
    Proto-Net ($\mathcal{L}_{id}$) &  0.398  &  0.387 & \bf  0.266  & \bf 0.285 & \bf 0.445 & 0.536 & 0.456 & 0.420 & 0.525 &  0.316 &  0.634 \\
    O-Proto ($\mathcal{L}^{ent}_{id} + \mathcal{L}^{hinge}_{id} + \mathcal{L}^{hinge}_{ood}$) & 0.348 & 0.375 & 0.411 & 0.409 & 0.631 & 0.643 & 0.404 & 0.390 & 0.482 & 0.373 & 0.683\\
    \hline
    \bf \underline{This study} & &&&& & &&  & &&\\
    ProtoInfoMax & 0.373 & 0.278 & 0.351 & 0.365 & 0.592 & 0.521 & 0.398 & \bf 0.368 &  0.398 & 0.256 & 0.549 \\
    ProtoInfoMax++  & \bf 0.335 & \bf 0.245 &  0.301  &   0.315 & 0.532 & \bf 0.469 & \bf 0.369 & 0.382 & \bf 0.388 & \bf 0.225 &  \bf 0.519 \\
    \hline
    \end{tabular}
}
    \caption{Performance for $K=100$ \footnotemark. The lower the better. Scores are based on top$-3$ the highest accuracy score for ID prediction (1-CER$^{\textnormal{id}}$) across meta-validation and meta-test episodes (epochs). For one class prediction of intent classification ($N=1$), EER and (1-CER$^{\textnormal{all}}$) are equal, and CER$^{\textnormal{id}}=1.0$ because the number of ID class within the subset is $1$. Evaluation for both $N=1$ and $N=2$ intent classification use the same model trained on $N=10, K=100$.
    }
    \label{tab:rsl_all}
\end{table*}

\subsection{Evaluation Metrics}
\label{sec:metric}

\paragraph{ID and OOD Detection Errors}

We use \textbf{(i) Equal Error Rate (EER)} for measuring error in predicting OOD; \textbf{(ii) Class Error Rate (CER$^{\textnormal{id}}$)} for measuring error in predicting ID examples; and \textbf{(iii) CER$^{\textnormal{all}}$} for measuring error in ID prediction given both ID and OOD subsets, following the previous work on OOD detection \cite{ryu-etal-2018-domain,tan-etal-2019-domain}. Except for CER$^{\textnormal{id}}$, metrics are calculated based on heuristically selected threshold value $\tau$. Given prediction outcomes with respect to decision whether examples are ID or OOD based on threshold $\tau$, the error rate scores are defined as:

\begin{equation}
    \textnormal{FAR} =\frac{\textnormal{FN}}{\small{\textnormal{\# OOD examples}}},
\end{equation}

\begin{equation}
    \textnormal{FRR} =\frac{\textnormal{FP}}{\small{\textnormal{\# ID examples}}},
\end{equation}

\begin{equation}
    \textnormal{EER} = \frac{1-(\textnormal{TP}+ \textnormal{TN}) }{ \small{\textnormal{\# Examples}}},
\end{equation}

\begin{equation}
    \textnormal{CER}^{\textnormal{id}} =\frac{\textnormal{TP}^{\textnormal{id}}}{\small{\textnormal{\# ID examples}}},
\end{equation}

\begin{equation}
    \textnormal{CER}^{\textnormal{all}} =\frac{\textnormal{TP}}{ \small{\textnormal{\# ID examples}}},
\end{equation}

\noindent where TN denotes correct OOD prediction based on threshold $\tau$. TP denotes correct ID prediction. FN measures OOD samples that are predicted as ID. FP measures ID samples that are predicted as OOD. TP$^{\textnormal{id}}$ is the number of correctly classified ID examples, excluding OOD samples.

\paragraph{Threshold score $\tau$} is calculated by heuristically searching a score conditioned by FRR and FAR metrics over sorted meta-test predictions \cite{ryu-etal-2018-domain,tan-etal-2019-domain}. That is, a score where the difference between False Acceptance Rate (FAR) and False Rejection Rate (FRR) has reached a minimum lower bound (FRR-FAR->0). Prior to the search, the initial threshold was defined as an average score of two prediction outcomes with the lowest scores. The final selected threshold $\tau$ is then used as a boundary score to distinguish between ID and OOD prediction. 



\footnotetext{Notice that our results (O-Proto performance) is different from those reported in \cite{tan-etal-2019-domain}. This might be due to different implementation frameworks: PyTorch vs. native Tensorflow; different hyper-parameters: we use $60 \times \#{\textnormal{Task}} \times 100$ batches due to our computational constraints vs. $5K \times \#{\textnormal{Task}} \times 100$ in \cite{tan-etal-2019-domain}; or different computing resources: GPU/CPU capacity used to train the models.}


\paragraph{Reliability Diagram} Reliability diagram \cite{niculescu2005predicting,Guo2017Calibration} depicts gaps between accuracy and model confidence. The larger the gap, the less calibrated the model is. That is, either the model is being underconfident or overconfident on estimating the winning predicted class labels. We use Expected \textbf{Calibration Error (ECE)} \cite{naeini2015obtaining,Guo2017Calibration} to summarize the difference in expectation between confidence and accuracy (gaps) across all bins. We use similarity score $d$ as a model confidence measure, following relevant work on distance-based prototypical learning \cite{Xing2020DistanceBased}.



\section{Results and Analysis}
\label{sec:results}

We demonstrate the effectiveness of our proposed methods (\textbf{ProtoInfoMax} and \textbf{ProtoInfoMax++}) on two benchmarks for OOD detection (Table~\ref{tab:rsl_all}). Notice that native Prototypical Networks (\textbf{Proto-Net}) performs reasonably well, specifically for ID prediction (see scores based on \textbf{CER}$^{id}$ and \textbf{CER}$^{all}$). However, this result can occur to models that always output predictions with a high score (e.g. high similarity score based on $d$ in the current work), regardless whether the prediction is correct. The insight into this overconfidence behaviour is provided in Section~\ref{sec:rel-id} and \ref{sec:rel-ood}.

\subsection{Performance in different K-shot}

Our \textbf{ProtoInfoMax} and \textbf{ProtoInfoMax++} also show a considerably consistent performance on meta-testing tasks under different $K$-shot values (Table~\ref{tab:eer_kshots_sent} and \ref{tab:eer_kshots_intent}), outperforming O-Proto.

\begin{table}[!ht]
\resizebox{.4\textwidth}{!}{
    \centering
    \begin{tabular}{l |ccc}
   \bf Model  & \bf EER & \bf CER$^{\textnormal{id}}$  & \bf CER$^{\textnormal{all}}$ \\
    \hline
    \bf \underline{K=1} &&&\\
    O-Proto   & 0.381  & 0.450  &  0.676  \\
    \bf ProtoInfoMax     & \bf 0.313 & 0.432 &  0.616  \\
    \bf ProtoInfoMax++     & 0.335  & \bf 0.430 & \bf 0.615  \\
    \hline
    \bf \underline{K=10} &&&\\
    O-Proto   & 0.311 & 0.425 &  0.606     \\
    \bf ProtoInfoMax     & 0.286 & 0.419 & 0.578   \\
    \bf ProtoInfoMax++     & \bf 0.254 & \bf 0.375 & \bf 0.537     \\
    \hline
    \bf \underline{K=100} &&&\\
    O-Proto   & 0.375 & 0.409 &  0.643  \\
    \bf ProtoInfoMax     & 0.278 & 0.365 &  0.521  \\
    \bf ProtoInfoMax++     & \bf 0.245 &  \bf 0.315 & \bf 0.469  \\
    \hline
    \end{tabular}
    }
    \caption{Performance under different $K$-shot values in sentiment classification ($N=2$). Scores are based on the highest accuracy ($1-\textnormal{CER}^{\textnormal{id}}$) on $\mathcal{T}^{\textnormal{test}}$.}
    \label{tab:eer_kshots_sent}
\end{table}

\begin{table}[!ht]
\resizebox{.4\textwidth}{!}{
    \centering
    \begin{tabular}{l |ccc}
   \bf Model  & \bf EER & \bf CER$^{\textnormal{id}}$  & \bf CER$^{\textnormal{all}}$ \\
    \hline
    \bf \underline{K=1} &&&\\
    O-Proto  & 0.515  & 0.391  &  0.698  \\
    \bf ProtoInfoMax     & 0.480 & 0.397 &  0.674  \\
    \bf ProtoInfoMax++     & \bf 0.452 & \bf 0.384 & \bf 0.638  \\
    \hline
    \bf \underline{K=10} &&&\\
    O-Proto   & 0.493 & 0.402 & 0.694  \\
    \bf ProtoInfoMax     &   0.451 & 0.400 & 0.686  \\
    \bf ProtoInfoMax++     & \bf 0.401 & \bf 0.329 & \bf 0.598 \\
    \hline
    \bf \underline{K=100} &&&\\
    O-Proto  & 0.482 &  0.373 & 0.683  \\
    \bf ProtoInfoMax &  0.398   & 0.256 & 0.549   \\
    \bf ProtoInfoMax++     & \bf 0.388 & \bf 0.225 &  \bf 0.519\\
    \hline
    \end{tabular}
    }
    \caption{Performance under different $K$-shot values in intent classification ($N=2$). }
    \label{tab:eer_kshots_intent}
\end{table}

\subsection{On Threshold Score, FAR, and FRR}



\begin{figure}[!ht]
    \centering
    \begin{subfigure}[ht]{.23\textwidth}
         \centering
         \includegraphics[width=\linewidth]{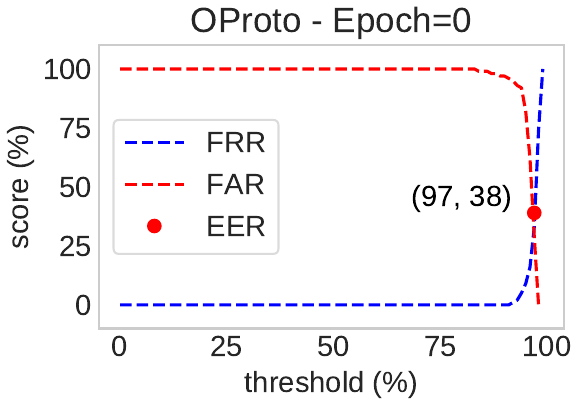}
       \end{subfigure}
    \begin{subfigure}[ht]{.23\textwidth}
         \centering
         \includegraphics[width=\linewidth]{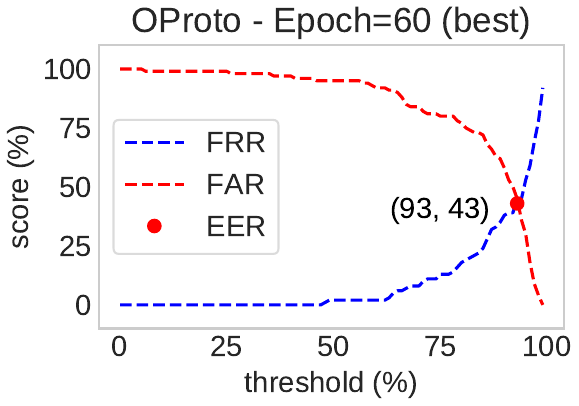}
       \end{subfigure}
    \begin{subfigure}[ht]{0.23\textwidth}
         \centering
         \includegraphics[width=\linewidth]{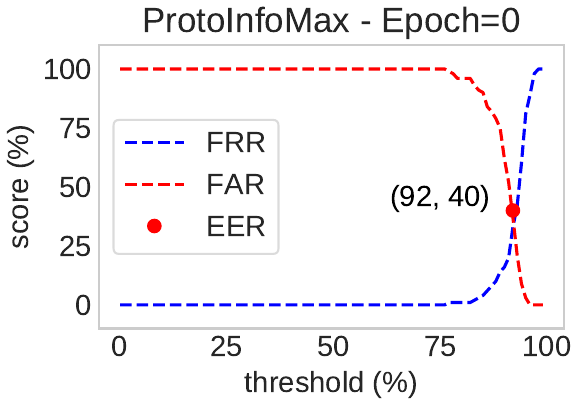}
       \end{subfigure}
    \begin{subfigure}[ht]{.23\textwidth}
         \centering
         \includegraphics[width=\linewidth]{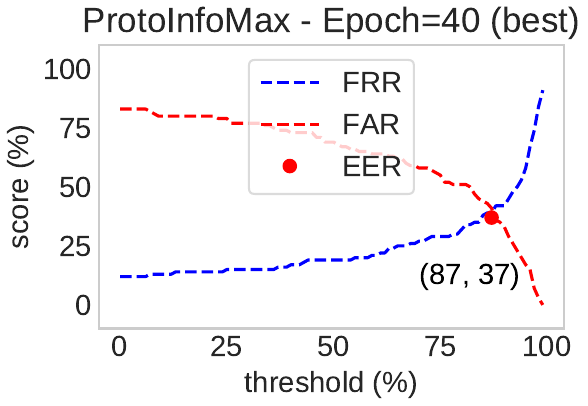}
       \end{subfigure}
    \begin{subfigure}[ht]{0.23\textwidth}
         \centering
         \includegraphics[width=\linewidth]{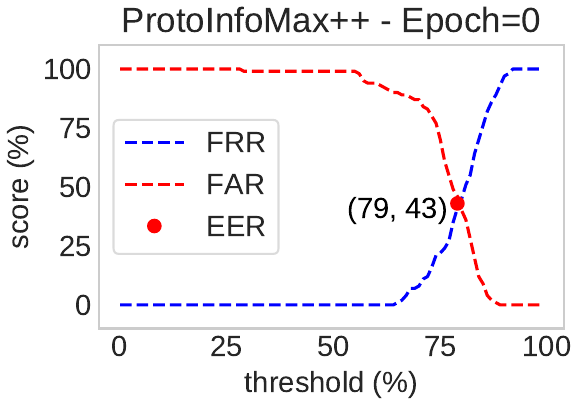}
       \end{subfigure}
    \begin{subfigure}[ht]{.23\textwidth}
         \centering
         \includegraphics[width=\linewidth]{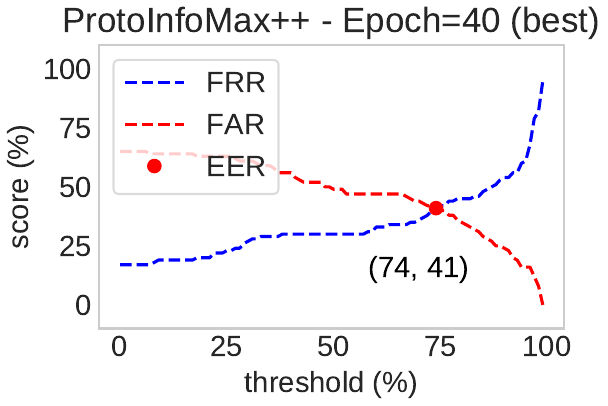}
       \end{subfigure}
    \caption{EER, FRR, FAR in intent classification meta-testing. Score $(\%)$ denotes proportion of samples that are either rejected (ID) or accepted (OOD) based on the selected threshold. To plot the above FAR and FRR, 200 prediction points corresponding to ID and OOD test samples were drawn randomly from 6 domains in $N=2$ meta-test episodes.}
    \label{fig:threshold-sc}
\end{figure}

We want to further inspect the reliability of model prediction. Figure~\ref{fig:threshold-sc} shows the selected threshold score across models in the intent classification task. It can be observed that \textbf{O-Proto} has a tendency to be overconfident, suggested by a considerably high threshold score ($\tau=0.97$ at epoch $0$ and $\tau=0.93$ at epoch $40$). Both \textbf{ProtoInfoMax} and \textbf{ProtoInfoMax++} are being less confident after several epochs, yielding lower thresholds ($\tau=0.87$ and $\tau=0.74$ respectively). Compared to O-Proto, \textbf{ProtoInfoMax++} converges faster in early episodes (epoch$=0$), yielding lower threshold score ($\tau=79$). 

We argue that one potential reason causing ProtoInfoMax++ to yield the lowest threshold score at epoch $0$ is due to an effective regularization via InfoMax objective, in addition to the use of multiple representations that promotes discriminative metric space. Early convergence pattern can also be observed on ProtoInfoMax, indicating that InfoMax objective is empirically shown to be beneficial on preventing overconfidence from early iteration. Notice that the gaps between FAR and FRR for both \textbf{ProtoInfoMax} and \textbf{ProtoInfoMax++} at epoch $=40$ are smaller. This indicates that both models underestimate ID and OOD samples, assigning them with low similarity scores ($d \leq 0.0$) with respect to ID class prototypes \footnote{We do not normalize $d, d \in[-1,1]$ here to inspect whether the model penalizes OOD samples severely with similarity score $d \leq 0.0$.}.

\subsection{Reliability in ID Prediction}
\label{sec:rel-id}

\begin{figure}[!ht]
    \centering
    \begin{subfigure}[ht]{.23\textwidth}
         \centering
         \includegraphics[width=\linewidth]{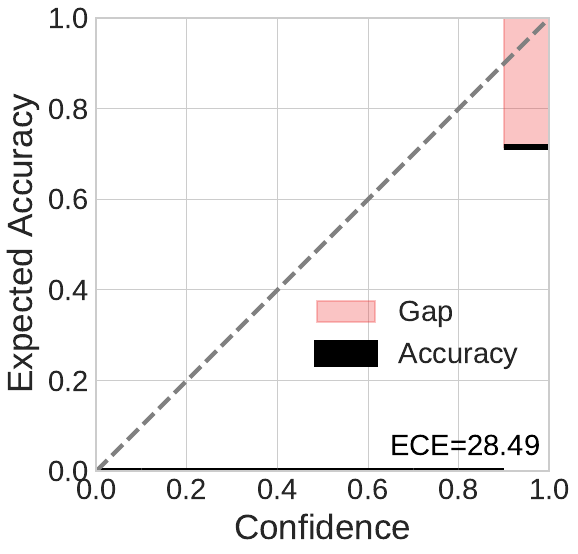}
    \caption{Proto-Net}
	\label{fig:id-reliability-a}
       \end{subfigure}
    \begin{subfigure}[ht]{.23\textwidth}
         \centering
         \includegraphics[width=\linewidth]{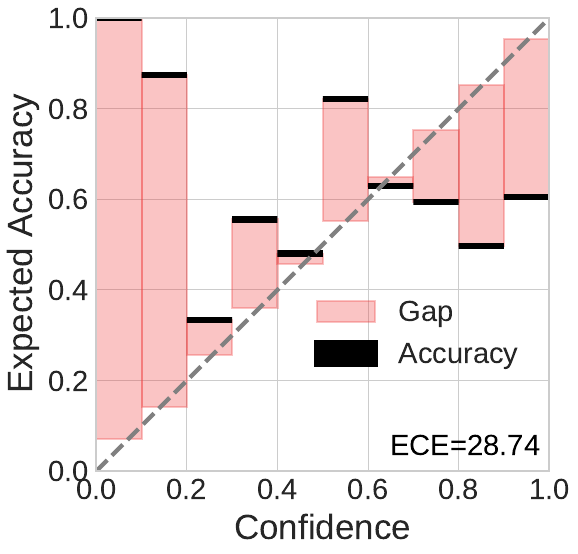}
    \caption{O-Proto}
	\label{fig:id-reliability-b}
       \end{subfigure}
    \begin{subfigure}[ht]{0.23\textwidth}
         \centering
         \includegraphics[width=\linewidth]{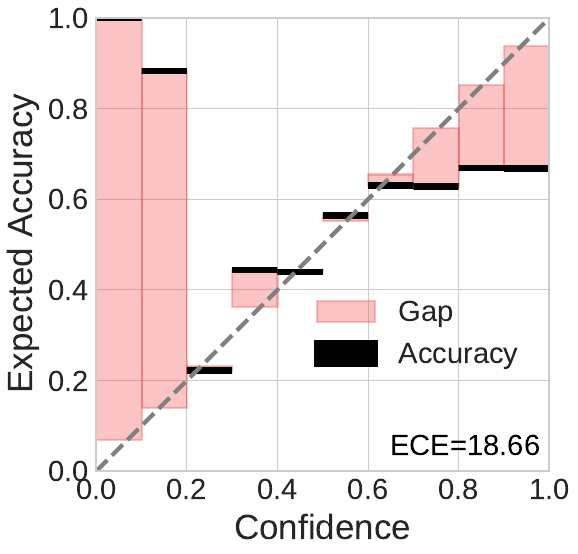}
	\caption{ProtoInfoMax}
	\label{fig:id-reliability-c}
       \end{subfigure}
    \begin{subfigure}[ht]{.23\textwidth}
         \centering
         \includegraphics[width=\linewidth]{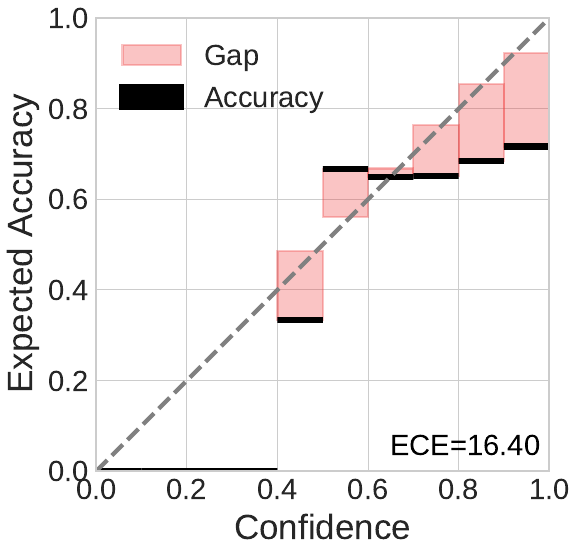}
    \caption{ProtoInfoMax++}
	\label{fig:id-reliability-d}
       \end{subfigure}
    \caption{Reliability Diagram for ID prediction. Confidence scores were taken from $\mathcal{T}^{test}$ in sentiment classification ($N=2, K=100$) based on the highest $1-\textnormal{CER}^{\textnormal{id}}$.}
    \label{fig:id-reliability}
\end{figure}

Figure~\ref{fig:id-reliability} compares the reliability of models in sentiment classification \footnote{Since OOD labels are unknown during training, this evaluation only includes the prediction outcomes from ID target queries as test samples.}. In general, all models in this study tend to be overconfident, suggesting that future work focusing on directly tackling and investigating such problem is essential \footnote{In current work, we abuse terminology of ``confidence score'' to refer to similarity score $d$, following relevant work on distance-based prototypical learning \cite{Xing2020DistanceBased}. }.





Compared to the baselines, our proposed \textbf{ProtoInfoMax} and \textbf{ProtoInfoMax++} are shown to be less prone to typical overconfidence problem with respect to smaller gaps between their confidence score and the prediction accuracy. \textbf{Proto-Net}, however, suffers greatly from such overconfidence problem. It can be observed that Proto-Net assigns high similarity scores ($d \geq 0.9$) for all prediction points (see accuracy is lower than confidence score in Figure~\ref{fig:id-reliability-a}). 

Our methods achieve the lowest ECE scores ($\textnormal{ECE}$ ProtoInfoMax $=18.66$ and $\textnormal{ECE} $ ProtoInfoMax++$ =16.40$), suggesting a better reliability with respect to smaller gaps between model's confidence score and prediction accuracy. \textbf{O-Proto} (Figure~\ref{fig:id-reliability-c}) and \textbf{ProtoInfoMax} (Figure~\ref{fig:id-reliability-b}) have both low confidence and overconfidence prediction. The models underestimate correct ID target queries (large gaps with high accuracy for $d \in (0.0, 0.2)$) and overestimate incorrect ID examples (large gaps with lower accuracy for $d \in (0.7, 1.0)$). 




\subsection{Reliability in OOD Prediction}
\label{sec:rel-ood}

\begin{figure}[!ht]
    \centering
    \begin{subfigure}[ht]{.23\textwidth}
         \centering
         \includegraphics[width=\linewidth]{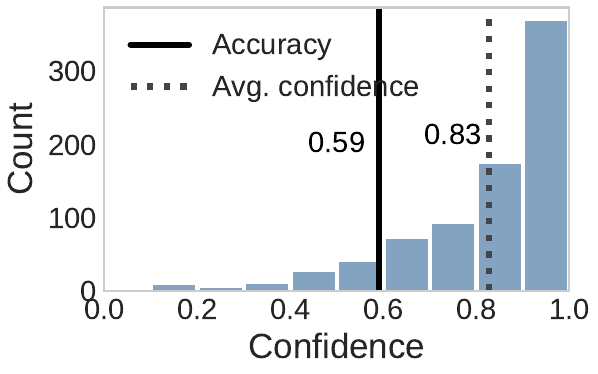}
    \caption{ID O-Proto}
	\label{fig:id-conf-a}
       \end{subfigure}
    \begin{subfigure}[ht]{0.23\textwidth}
         \centering
         \includegraphics[width=\linewidth]{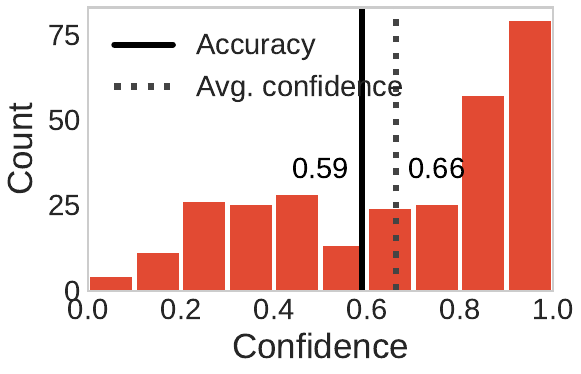}
	\caption{OOD O-Proto}
	\label{fig:ood-conf-a}
       \end{subfigure}
    \begin{subfigure}[ht]{.23\textwidth}
         \centering
         \includegraphics[width=\linewidth]{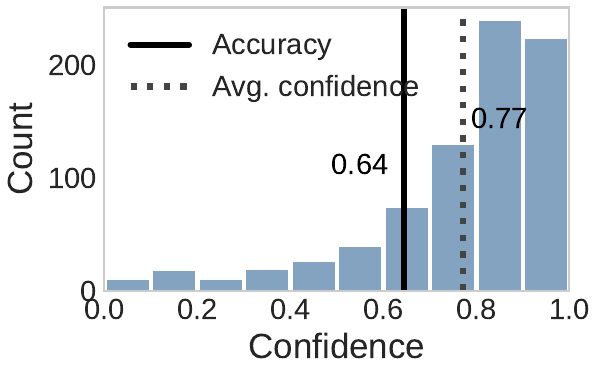}
    \caption{ID ProtoInfoMax}
	\label{fig:id-conf-b}
       \end{subfigure}
    \begin{subfigure}[ht]{0.23\textwidth}
         \centering
         \includegraphics[width=\linewidth]{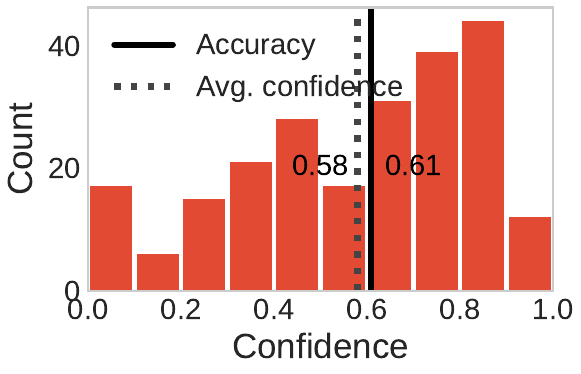}
	\caption{OOD ProtoInfoMax}
	\label{fig:ood-conf-b}
       \end{subfigure}
    \begin{subfigure}[ht]{.23\textwidth}
         \centering
         \includegraphics[width=\linewidth]{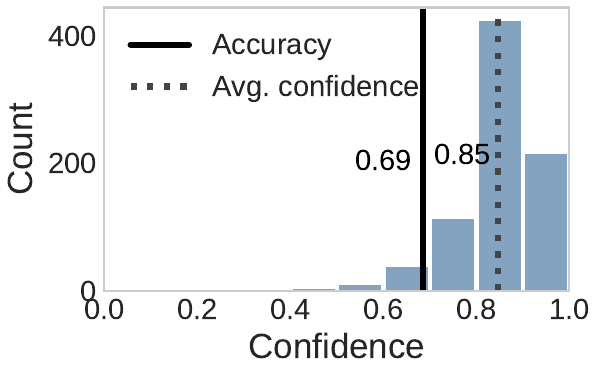}
    \caption{ID ProtoInfoMax++}
	\label{fig:id-conf-c}
       \end{subfigure}
    \begin{subfigure}[ht]{0.23\textwidth}
         \centering
         \includegraphics[width=\linewidth]{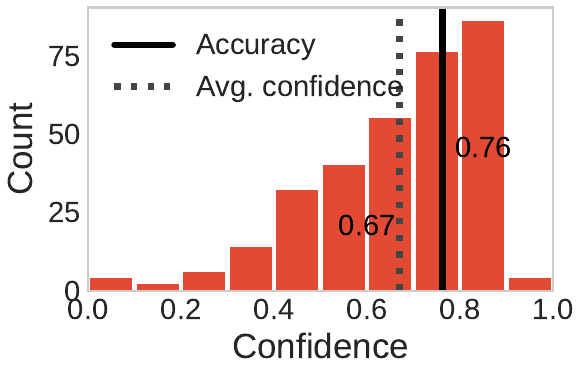}
	\caption{OOD ProtoInfoMax++}
	\label{fig:ood-conf-c}
       \end{subfigure}
    \caption{Confidence histogram for ID and OOD prediction. Confidence scores were taken from $\mathcal{T}^{test}$ in sentiment classification ($N=2, K=100$). Note that the value of accuracy and average confidence here are not as precise as 1-CER$^{\textnormal{id}}$ score, since they were averaged across normalized bin scores.}
    \label{fig:id-ood-conf}
\end{figure}

The reliability based on confidence histogram for ID and OOD prediction is provided in Figure~\ref{fig:id-ood-conf} \footnote{Since OOD labels are unavailable during training, the reliability diagram is not applicable for evaluating OOD prediction.}. In general, all models overestimate their prediction given ID target queries (see that the average confidence is higher than accuracy in Figure~\ref{fig:id-conf-a}, \ref{fig:id-conf-b}, \ref{fig:id-conf-c}). However, compared to \textbf{O-Proto}, our \textbf{ProtoInfoMax} and \textbf{ProtoInfoMax++} have  a higher accuracy in ID classification tasks given their reasonably high confidence. Notice that for the ID prediction task, \textbf{ProtoInfoMax++} is more confident than the other two models ($d \in (0.4, 1.0)$ in Figure~\ref{fig:id-conf-c}). 

For OOD detection \footnote{Here, we view the task as one class OOD prediction where test samples contain OOD target queries only. Values below threshold $\tau$ are classified as OOD. Values above threshold are classified as ID.}, our \textbf{ProtoInfoMax} and \textbf{ProtoInfoMax++} are shown to be less prone to overconfidence problem than \textbf{O-Proto}. See that the average confidence scores of both models are lower than their prediction accuracy (ProtoInfoMax avg. confidence: $0.58$ in Figure~\ref{fig:ood-conf-b} and ProtoInfoMax++: $0.67$ in Figure~\ref{fig:ood-conf-c}). In contrast, the average confidence score of \textbf{O-Proto} is higher than its prediction accuracy (Avg. Confidence $=0.66$, Accuracy$=0.59$ in Figure~\ref{fig:ood-conf-a}), indicating the model prediction with an overconfidence issue.

\section{Conclusion}
\label{sec:conclude}

Simultaneously learning In-Domain (ID) text classification and Out-of-Domain (OOD) detection under low resource constraints is realistic but under-explored. In this study, we aim at effectively and reliably learning zero-shot Out-of-Domain detection via Mutual Information Maximization (InfoMax) objective. Although we do not specifically tackle overconfidence problem of Neural Networks by calibrating models during training and evaluation stage in the current OOD detection task, we observe that the proposed ProtoInfoMax and ProtoInfoMax++ are less prone to such typical overconfidence problem compared to existing approaches. Overall, we improve performance of existing approaches up to 20\% for OOD detection in low resource text classification. 

\section*{Acknowledgment}
This research is supported by Indonesian Endowment Fund for Education (LPDP) Scholarship under Beasiswa Pendidikan Indonesia (BPI) -- ID Number 0003194/SC/D/9/LPDP2016. The content of the information does not necessarily reflect the position or the policy of the Government, and no official endorsement should be inferred.

\bibliography{anthology,custom}
\bibliographystyle{acl_natbib}


\end{document}